\documentclass{article}
\usepackage{iclr2027_conference,times}

\usepackage{amsmath,amsfonts,bm}

\def\eqref#1{equation~\ref{#1}}

\def\1{\bm{1}}

\DeclareMathAlphabet{\mathsfit}{\encodingdefault}{\sfdefault}{m}{sl}
\SetMathAlphabet{\mathsfit}{bold}{\encodingdefault}{\sfdefault}{bx}{n}

\usepackage{booktabs}
\usepackage{float}
\usepackage{graphicx}
\usepackage{hyperref}
\usepackage{url}

\title{Transfer Learning for Edge Classification on Dynamic Text-Attributed Graphs}

\author{Tyler Bonnet \& Marek Rei \\
Imperial College London \\
London, United Kingdom \\
\texttt{\{t.bonnet24, marek.rei\}@imperial.ac.uk}
}

\iclrfinalcopy

\begin{document}

\maketitle

\begin{abstract}
Learning transferable representations for dynamic text-attributed graphs (DyTAGs) requires models to capture underlying interaction dynamics that persist across domains. However, existing methods tend to overfit to domain-specific structural, temporal, and semantic patterns, limiting edge classification performance under distribution shifts. To expose and address this, we formally establish a leave-one-domain-out (LODO) transfer learning protocol for edge classification on DyTAGs. Under this protocol, we demonstrate that state-of-the-art self-supervised methods for dynamic graph learning perform poorly when transferred to unseen domains. Strikingly, existing methods underperform a structurally and temporally unaware Bag of Events (BoE) model we introduce, which inputs only unordered sequences of node and edge text features. Proceeding from the BoE, we propose Spatio-Temporal Semantic Alignment (STSA), which integrates a spatio-temporal encoder that fuses representations of time deltas and node occurrence frequencies into a unified manifold. STSA is trained with a Contrastive Semantic Forecasting objective, which anchors edge representations to a multi-domain textual latent space initialized by a pretrained language model, providing a robust prior that outperforms BoE and all existing methods we evaluate.
\end{abstract}

\section{Introduction}
\label{intro}
Dynamic text-attributed graphs (DyTAGs) serve as useful representations for complex systems where inherently linguistic or semantically annotated interactions between entities evolve over time, appearing frequently in domains such as e-commerce, temporal knowledge graphs, and communication networks. Standard frameworks model interactions through either discrete snapshot-based methods \citep{sankar2020dysat, pareja2020evolvegcn} or continuous-time methods that process data as a streaming sequence \citep{kumar2019jodie, trivedi2019dyrep, rossi2020temporal}. Transformer architectures \citep{wang2021tcl, yu2023dygformer} advance this sequential paradigm, using self-attention over historical events to capture complex dependencies without relying on the restrictive compression of recurrent memory states. Such models were primarily designed for link prediction, a task where training labels are inherently abundant because they are derived from the topology of the graph. In contrast, the task of edge classification involves predicting the nature of an interaction, which frequently relies on scarce extrinsic annotations. Training highly parameterized models on limited data often causes them to overfit or collapse to majority class predictors. Self-supervised learning on unlabeled data becomes essential to train more robust models \citep{xu2023cldg, zhu2024idol, liu2025dygmae, gao2025dvgmae, bonnet2026dygnrole}. Capturing transferable interaction patterns across diverse domains improves generalizability further and allows developers to avoid the computational costs of pretraining a new model for every domain \citep{wang2023evolunet, shamsi2025mint, huang2025crosslink, qiao2025gcal, chatterjee2025tlp}.

However, the field has yet to formally address cross-domain transfer learning for edge classification, as current frameworks rely on problem formulations limited to in-domain distributions. To address this, our \textbf{first contribution} is to formally establish a leave-one-domain-out (LODO) transfer learning protocol for edge classification on DyTAGs, which requires models to predict the labels of future interactions in an unseen target domain after self-supervised pretraining on a set of source domains.

Our \textbf{second contribution} establishes a two-part diagnostic framework to expose critical vulnerabilities of existing approaches to modeling time and local structure. First, we conduct a representational analysis, demonstrating that when temporal and structural features are processed without non-linear fusion, their signals fail to properly integrate. Upon cross-domain transfer, this causes the representations to fragment, restricting the effective capacity of the model. Second, we introduce a Bag of Events (BoE) baseline. Pretrained solely on unordered sequences of node and edge text features, BoE serves as a control to determine whether structural or temporal inputs are sources of performance degradation. Strikingly, all existing methods we evaluate underperform BoE on average.

To address these vulnerabilities, our \textbf{third contribution} proposes Spatio-Temporal Semantic Alignment (STSA), which integrates a spatio-temporal encoder (STE) that fuses representations of time deltas and node frequencies into a unified manifold to resolve the fragmentation and expand representational capacity. We propose Contrastive Semantic Forecasting (CSF) to pretrain STSA. CSF allows more organizational freedom for representations than point-for-point reconstruction while anchoring the model to the multi-domain latent space of a pretrained language model (PLM).

Our \textbf{fourth contribution} lies in extensive empirical evaluations on the real-world datasets of the Dynamic Text-Attributed Graph Benchmark (DTGB) \citep{zhang2024dtgb}. STSA consistently achieves statistically significant performance gains over BoE and all other existing methods across all DTGB domains. Furthermore, we isolate the quality of the pretrained representations of frozen encoders with a $K$-shot evaluation, demonstrating that STSA outperforms existing methods across all values of $K$. Finally, an ablation study confirms the necessity of STE and the superiority of CSF over InfoNCE \citep{oord2018infonce} and Scaled Cosine Error (SCE) \citep{hou2022graphmae} losses.

\section{Related Work}
\label{related_work}

Research on pretraining generalizable dynamic graph models has largely proceeded along two parallel tracks: self-supervised learning within domains and transfer learning across domains. Importantly, frameworks in both tracks operate with pretraining steps that are decoupled from downstream evaluation. This distinguishes them from recent LLM-driven models for DyTAGs \citep{zhang2025cross, roy2025lkd4dytag, xu2026moment, wang2026dygrasp}, which use joint-optimization paradigms that rely heavily on downstream task labels and cannot be easily isolated for standalone pretraining.

In the self-supervised track, pretraining strategies generally divide into generative reconstruction and contrastive alignment. Generative architectures formulate pretraining as a missing data recovery task by using masking strategies to reconstruct graph topology or temporal features \citep{liu2025dygmae, gao2025dvgmae}. Conversely, contrastive methods optimize for mutual information by aligning varied structural or temporal views of the graph. This is often done by tracking topological changes or temporal neighborhood alignments \citep{xu2023cldg, zhu2024idol}. More recently, models have used Transformer architectures with specialized embedding tables for directional role alignment \citep{bonnet2026dygnrole}. While these methods learn generalizable representations, they are typically evaluated within-domain and target intra-graph distribution shifts rather than bridging new domains.

To address domain shifts, the transfer learning track attempts to capture domain-agnostic interaction patterns. One line of research scales pretraining across multiple graphs to prevent encoders from memorizing graph-specific biases via training order shuffling or temporal normalization \citep{shamsi2025mint, huang2025crosslink}. Another relies on algorithmic adaptation to align distributions. This includes adversarial frameworks extracting domain-invariant features \citep{wang2023evolunet}, continual learning strategies condensing synthetic memories \citep{qiao2025gcal}, and structural mapping approaches initializing target memory states from topological statistics \citep{chatterjee2025tlp}.

Although progress has been made in transfer learning for link prediction, node classification, and graph property prediction, none of the reviewed works address the significant challenges of transferring specifically for edge classification. STSA departs from the reviewed methods to tackle this directly. While other Transformer-based models exist, they introduce additional parameters for explicit role modeling, exacerbating overfitting \citep{bonnet2026dygnrole}, discard all node and edge attributes \citep{huang2025crosslink}, or enforce domain invariance through adversarial alignment \citep{wang2023evolunet}. Moreover, none of the methods fuse temporal and structural representations to mitigate overfitting to these modalities (STE) or align with a multi-domain PLM embedding space to improve transferability (CSF).

\section{Methodology}
\label{method}

This section details the formal setup of the leave-one-domain-out (LODO) transfer learning protocol for dynamic edge classification, our proposed Bag of Events (BoE) baseline, and our Spatio-Temporal Semantic Alignment (STSA) framework.

\subsection{Preliminaries}
\label{method:problem}

We represent a dynamic text-attributed graph as $\mathcal{G} = (\mathcal{V}, \mathcal{E}, \mathcal{T})$, where $\mathcal{V}$ is the node set, $\mathcal{E}$ is the edge set, and $\mathcal{T}$ is the set of timestamps. Each interaction $\epsilon \in \mathcal{E}$ is a chronologically ordered tuple $\epsilon = (u, v, t)$ for $u, v \in \mathcal{V}$ and $t \in \mathcal{T}$. Nodes and interactions are associated with text attributes. For any given timestamp $t$, the temporal history of the graph is the set of all prior interactions $\mathcal{E}_{< t} = \{\epsilon \in \mathcal{E} \mid t_\epsilon < t\}$. Interactions $\mathcal{E}_{label} \subset \mathcal{E}$ are provided with a ground-truth label $y_\epsilon \in \mathcal{Y}$.

\subsection{The Leave-One-Domain-Out (LODO) Protocol}
\label{method:lodo}

To rigorously investigate transfer learning for edge classification on DyTAGs, we formally establish the leave-one-domain-out (LODO) protocol. Under this protocol, a domain can consist of one or more dynamic graphs. We define a set of source domains $\mathcal{D}$ for pretraining, which collectively contain a set of source graphs $\mathcal{S} = \{\mathcal{G}_1, \ldots, \mathcal{G}_K\}$. For evaluation, we hold out an entirely unseen target domain and evaluate on a constituent target graph $\mathcal{G}_{target}$. The learning process is divided into two phases. First, we learn an encoder $h_{\theta}$ over $\mathcal{S}$ without using edge labels. Second, we introduce a task-specific classification head $g_{\phi}$ for the dynamic edge classification task on $\mathcal{G}_{target}$. The model is then adapted using the labeled subset $\mathcal{E}_{label} \subset \mathcal{E}_{target}$, accommodating downstream paradigms such as end-to-end finetuning or frozen-encoder linear probing. For a new interaction between $u$ and $v$ arriving at time $t$ in $\mathcal{G}_{target}$, the composite function may use any information about the query nodes or interaction time $t$, along with the available history $\mathcal{E}_{target < t}$, to predict the edge label $\hat{y}_\epsilon$. All information about the query edge itself must be masked from the input:

\begin{equation}
    \label{eq:lodo}
    \hat{y}_\epsilon = g_{\phi}(h_{\theta}(u, v, t, \mathcal{E}_{target < t})).
\end{equation}

\subsection{Bag of Events (BoE)}
\label{method:boe}

The BoE model serves as a control for structural and temporal overfitting in existing models (illustrated in Appendix~\ref{app:boe_architecture}). Conceptually, it treats the interaction history of a node as an unordered sequence of text-attributed events, disregarding topology and timestamps. 

For a query interaction between nodes $u$ and $v$ at time $t$, we retrieve their respective historical one-hop interactions prior to $t$, truncated or padded to a fixed length $k$. Prepending $u$ and $v$ to their histories produces distinct event sequences $S_u$ and $S_v$. Textual features are the sole input to BoE. Node and edge text embeddings are precomputed using Qwen3-Embedding-4B \citep{zhang2025qwen3} and held static. For each event $i \in S_u$ (and symmetrically for $S_v$), let $\mathbf{n}_i \in \mathbb{R}^{d_{node}}$ and $\mathbf{e}_i \in \mathbb{R}^{d_{edge}}$ denote the node and edge text embeddings, respectively. To prevent target leakage, the text embedding of the query edge itself is masked. The embeddings are independently projected via weight matrices $\mathbf{W}_n$ and $\mathbf{W}_e$, and concatenated to form the event representation $\mathbf{x}_i = [\mathbf{W}_n \mathbf{n}_i \parallel \mathbf{W}_e \mathbf{e}_i] \in \mathbb{R}^{2d_c}$, where $\parallel$ denotes concatenation and $d_c = 96$. The sequence of these representations forms the input matrix $\mathbf{X}_u \in \mathbb{R}^{k \times 192}$ for $S_u$ (and $\mathbf{X}_v$ for $S_v$), which is encoded by a multi-layer Transformer \citep{vaswani2017attention} using a phase-dependent processing strategy.

During self-supervised pretraining, $\mathbf{X}_u$ and $\mathbf{X}_v$ are processed independently using shared weights to prevent trivial cross-attention shortcuts. To produce single vector representations, we apply masked mean pooling over the sequence dimension (excluding padding tokens), followed by $L_2$-normalization to yield final representations $\mathbf{z}_u$ and $\mathbf{z}_v$. BoE optimizes a symmetric InfoNCE loss \citep{oord2018infonce} with temperature $\tau=0.07$ across a minibatch to align $\mathbf{z}_u$ and $\mathbf{z}_v$ (formulation provided in Appendix~\ref{app:pretraining_objectives}). 

Conversely, during downstream finetuning for edge classification, the model must capture complex dependencies between the interacting neighborhoods. The sequences are therefore concatenated into a joint input $\mathbf{X}_{uv} = [\mathbf{X}_u \parallel \mathbf{X}_v]$ and processed jointly. After self-attention, the output sequence is split back into updated representations for $S_u$ and $S_v$. Following masked mean pooling, the resulting $\mathbf{z}_u$ and $\mathbf{z}_v$ are concatenated and passed to a linear classification head.

\subsection{Spatio-Temporal Semantic Alignment (STSA)}

\begin{figure}[t!]
    \centering
    \includegraphics[width=\textwidth]{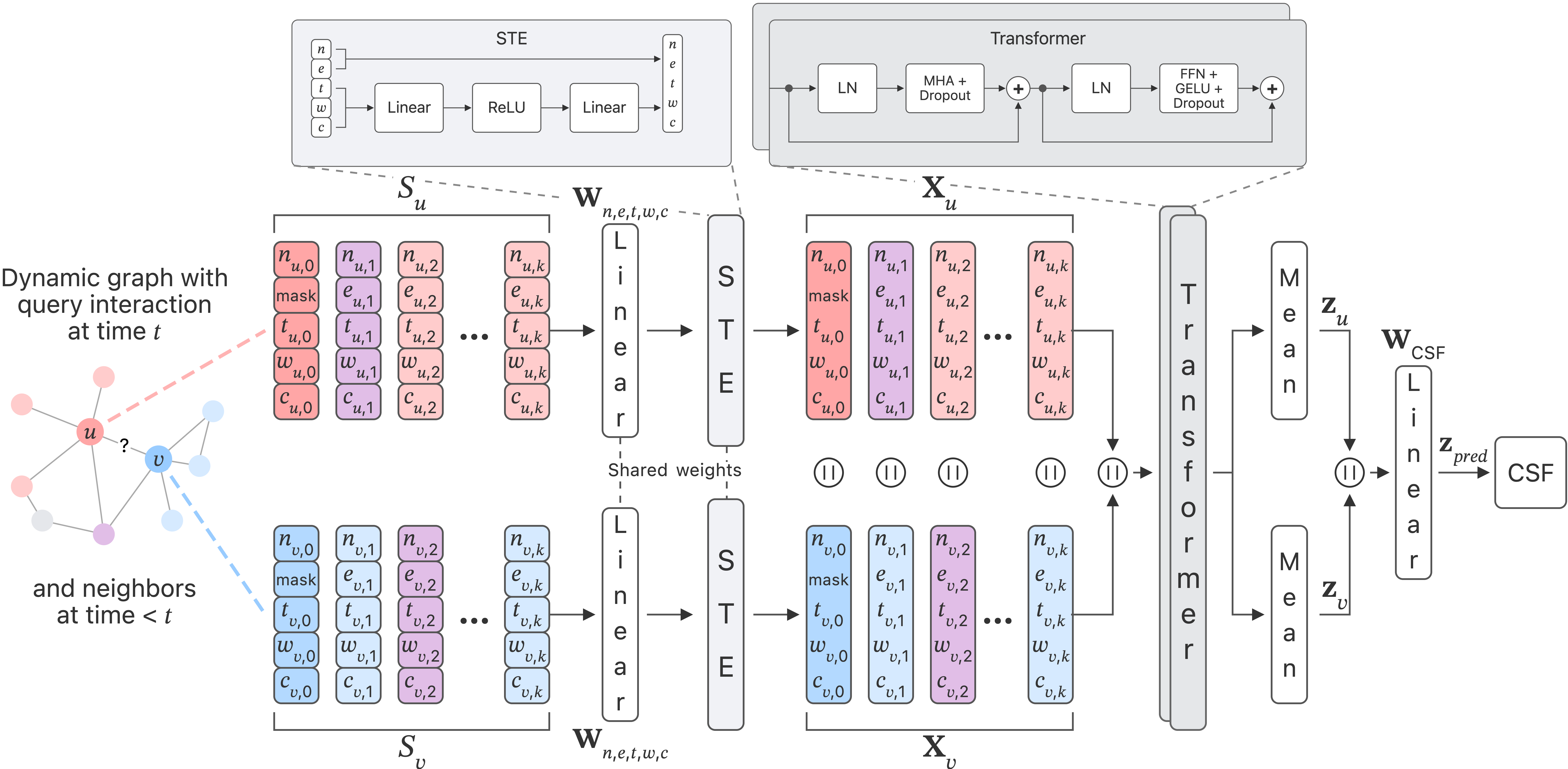}
    \caption{\textbf{Overview of the Spatio-Temporal Semantic Alignment (STSA) Framework.} \textbf{(Left)} For a query interaction between nodes $u$ and $v$ at time $t$, historical one-hop neighbors are retrieved. $u$ and $v$ combined with their respective neighbors form node sequences $S_u$ and $S_v$. For each event $i$ in these sequences, its node text embedding $\mathbf{n}_i$, edge text embedding $\mathbf{e}_i$, interaction time delta $\mathbf{t}_i$, within-sequence node frequency embedding $\mathbf{w}_i$, and cross-sequence node frequency embedding $\mathbf{c}_i$ are extracted, and the edge text embedding of the query nodes is masked. \textbf{(Center)} All five extracted features $\mathbf{n}_i$, $\mathbf{e}_i$, $\mathbf{t}_i$, $\mathbf{w}_i$, $\mathbf{c}_i$ are first independently projected via separate linear layers $\mathbf{W}_{n,e,t,w,c}$. The spatio-temporal encoder (STE) then fuses the projected temporal and structural features $\mathbf{t}_i$, $\mathbf{w}_i$, and $\mathbf{c}_i$ into a unified manifold. This fused spatio-temporal representation is concatenated with the projected node and edge text embeddings to form a single feature vector representing each event in $S_u$ and $S_v$. \textbf{(Right)} These event sequences, corresponding to input matrices $\mathbf{X}_u$ and $\mathbf{X}_v$, are concatenated on the sequence dimension and jointly processed by a multi-layer Transformer encoder. After the output sequence is split back into distinct representations for $S_u$ and $S_v$, mean pooling, with a mask to exclude padding tokens, is applied to form the final node representations $\mathbf{z}_u$ and $\mathbf{z}_v$. These representations are passed through a linear task head $\mathbf{W}_{\mathrm{CSF}}$ to yield $\mathbf{z}_{pred}$, which is optimized via Contrastive Semantic Forecasting (CSF) to predict the masked query edge, anchoring the model to the multi-domain textual latent space.}
    \label{fig:stsa_architecture}
\end{figure}

In this section, we introduce STSA (illustrated in Figure~\ref{fig:stsa_architecture}), a novel framework targeted at resolving structural, temporal, and semantic overfitting during transfer by integrating a spatio-temporal encoder (STE) and Contrastive Semantic Forecasting (CSF) into the architectural backbone of BoE.

\subsubsection{Spatio-Temporal Encoder (STE)}
\label{method:ste}

A fundamental challenge in dynamic graph learning is the modality gap between structure and time. As we demonstrate with a representational analysis in Section~\ref{sec:rep_analysis}, prior to Transformer encoding, these signals are severely fragmented and have restricted effective capacity under cross-domain transfer. To prevent this, STSA non-linearly fuses structural and temporal representations with an STE before sequence encoding. 

For each event $i \in S_u$ (and symmetrically for $S_v$), let $\Delta t$ be the time elapsed since the query interaction. We encode $\Delta t$ via a Fourier time encoder \citep{xu2020tgat} to produce $\mathbf{t}_i \in \mathbb{R}^{32}$. We map the occurrence count of the node within its own sequence and within the sequence of the paired node to embeddings $\mathbf{w}_i, \mathbf{c}_i \in \mathbb{R}^{32}$ \citep{bonnet2026dygnrole}, capturing local structure.

After projection via separate linear layers, the three 32-dimensional features representing $\Delta t$, within-sequence frequency, and cross-sequence frequency are concatenated to form a single 96-dimensional vector for each event. These vectors are processed by the STE, which is an MLP with a 256-dimensional hidden layer and a ReLU activation, to project the raw features into a unified 96-dimensional spatio-temporal latent space. This spatio-temporal representation is concatenated with the 192-dimensional semantic feature (detailed in Section~\ref{method:boe}) to form a 288-dimensional representation for each event in $S_{u}$ and $S_{v}$, yielding $\mathbf{X}_u$ and $\mathbf{X}_v$.

\subsubsection{Contrastive Semantic Forecasting (CSF)}
\label{method:csf}

For STSA pretraining, we propose CSF, an objective designed to predict the semantics of the interaction between $u$ and $v$ at time $t$. $\mathbf{X}_u$ and $\mathbf{X}_v$ are concatenated and processed jointly by the Transformer encoder. By forecasting the masked query edge rather than contrasting the nodes, CSF allows STSA to safely use joint sequence processing without target leakage. Following joint self-attention, the output sequence is split back into the distinct, updated representations for $S_{u}$ and $S_{v}$. Finally, to produce single vector representations for $u$ and $v$, we apply mean pooling over the sequence dimension excluding padding tokens from the calculation to yield the final representations $\mathbf{z}_{u}$ and $\mathbf{z}_{v}$.

To generate the predicted semantic representation $\mathbf{z}_{pred}$, the final node representations $\mathbf{z}_u$ and $\mathbf{z}_v$ from the Transformer are concatenated and fed through a dedicated CSF linear task head, which projects the joint state to 256 dimensions via $\mathbf{z}_{pred} = \mathbf{W}_{\mathrm{CSF}} [\mathbf{z}_u \parallel \mathbf{z}_v] + \mathbf{b}_{\mathrm{CSF}}$, where $\mathbf{W}_{\mathrm{CSF}}$ and $\mathbf{b}_{\mathrm{CSF}}$ are learnable parameters. For the ground-truth target $\mathbf{z}_{target}$, we use precomputed Qwen3 edge text embeddings of the interactions truncated to 256 dimensions. Because Qwen3-Embedding supports matryoshka representation learning \citep{kusupati2022matryoshka}, it enables this truncation while retaining the vast majority of the most important semantic information.

The model is optimized using an InfoNCE approach to align the predicted and target representations. For a minibatch of size $N$, the loss is formulated as:

\begin{equation}
    \label{eq:csf}
    \mathcal{L}_{\mathrm{CSF}} = -\frac{1}{N} \sum_{i=1}^{N} \log \frac{\exp\!\left(\frac{\mathbf{z}_{pred}^{(i)} \cdot \mathbf{z}_{target}^{(i)}}{\tau}\right)}{\sum_{j=1}^{N} \exp\!\left(\frac{\mathbf{z}_{pred}^{(i)} \cdot \mathbf{z}_{target}^{(j)}}{\tau}\right)},
\end{equation}

where $\tau$ is a temperature parameter set to 0.07, following standard practice in contrastive learning.

Instead of forcing point-for-point reconstruction with a mean squared error or SCE \citep{hou2022graphmae} loss, CSF allows more organizational freedom for the latent space by having the model produce a representation more similar to the true edge text embedding than to all in-batch negatives. Consequently, this objective incentivizes the model to learn the semantic trajectory of the interaction dynamics of the graph.

\section{Experiments}
\label{experiments}

In this section, we first conduct a representational analysis to demonstrate the mechanistic basis of the spatio-temporal encoder (STE), revealing how it non-linearly fuses structural and temporal signals to mitigate representational fragmentation and expand effective capacity under cross-domain transfer. We then evaluate BoE and STSA alongside existing self-supervised and transfer learning baselines on dynamic edge classification under the LODO protocol. We report performance under both end-to-end finetuning and a parametric $K$-shot linear probe that assesses the inherent quality of the transferred representations. Finally, we conduct an ablation study to isolate the contributions of STE and CSF.

\subsection{Spatio-Temporal Representational Analysis}
\label{sec:rep_analysis}

\begin{figure}[t]
    \centering
    \includegraphics[width=\textwidth]{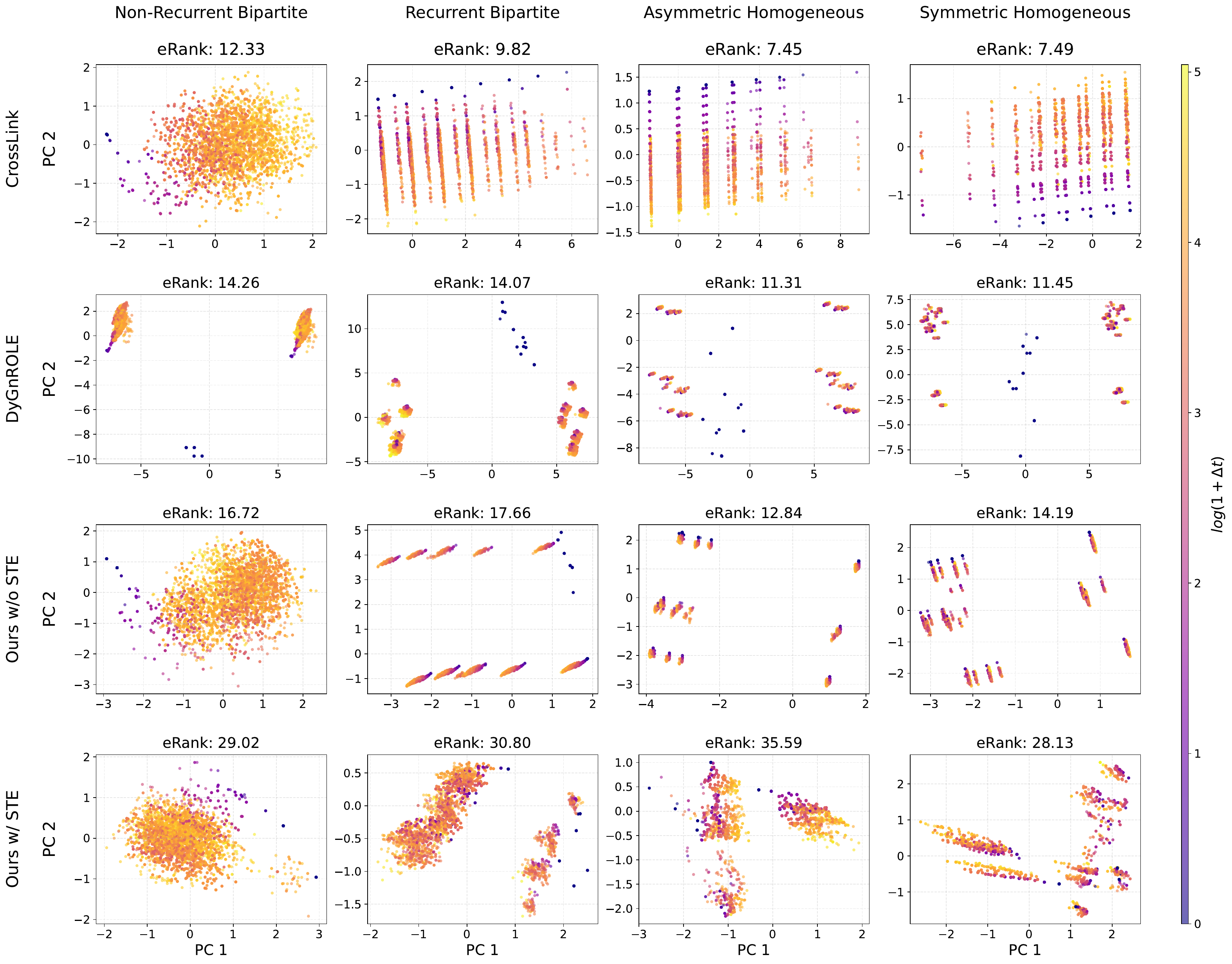}
    \caption{\textbf{Analysis of Spatio-Temporal Representations.} Principal Component Analysis (PCA) of spatio-temporal representations colored by $\log(1 + \Delta t)$ across four synthetic graph topologies. \textbf{Top 3}: Under domain transfer, concatenated time and node frequency features suffer from representational fragmentation and a restricted Effective Rank (eRank). \textbf{Bottom}: STE fuses the temporal and structural features into a unified manifold and substantially improves eRank.}
    \label{fig:ste_geometry_time_delta}
\end{figure}

To demonstrate the mechanistic necessity of fusing temporal and structural features via STE, we conduct a representational analysis (Figure~\ref{fig:ste_geometry_time_delta}). We isolate these modalities by generating text-free dynamic graphs across four archetypal topologies (detailed in Appendix~\ref{app:synthetic_generation}).

Using the BoE/STSA backbone with independent Transformer processing (Section~\ref{method:boe}), we train two text-free variants (with and without STE) with InfoNCE under the LODO protocol. We also train two event-based Transformer architectures, DyGnROLE and CrossLink, on their native objectives. After validation early stopping with a patience of 10, we extract target-domain features immediately prior to Transformer encoding and evaluate the manifold geometry via Principal Component Analysis (PCA) and dimensional health via Effective Rank (eRank) \citep{roy2007effective}.

As shown in Figure~\ref{fig:ste_geometry_time_delta} (top 3), the time and node frequency features of DyGnROLE, CrossLink, and Ours w/o STE suffer severe representational fragmentation and a restricted eRank of 7.45--17.66 out of 96. Without fusion, a Transformer must expend self-attention capacity bridging these gaps. STE resolves this via non-linear fusion into a more expressive manifold (bottom) that approximately doubles eRank on average. The clusters are artifacts of the within- and cross-sequence frequency representations of the models. We include another version of this figure coloring points by count combinations instead of $\log(1 + \Delta t)$ in Appendix~\ref{app:ste_geometry_node_frequency}. The separation of DyGnROLE representations for the Non-Recurrent Bipartite domain is a result of its additive role-semantic positional encodings, allowing the model to differentiate source-sequence nodes from destination-sequence nodes.

The Non-Recurrent Bipartite domain highlights a particular behavior of STE. In topologies with very little recurrence, within-sequence and cross-sequence node frequencies are almost always 1 and 0 (98.52\% of nodes in our generated graph), respectively, meaning the spatio-temporal signal is largely temporal. Although PCA visualizes the representations of most models in this domain as visually similar linear gradients, the eRank with STE confirms a large capacity expansion from a baseline maximum of 16.72 to 29.02.

\subsection{Benchmark Evaluation Setup}
\label{sec:setup}

\textbf{Datasets.} We evaluate on eight datasets from DTGB \citep{zhang2024dtgb}, spanning four diverse domains: e-commerce reviews (Amazon, Google, Yelp), corporate communication (Enron), technical Q\&A forums (Stack E, Stack U), and political events (GDELT, ICEWS). These domains exhibit high topological, temporal, and linguistic variance, alongside widely varied dynamic edge classification tasks, making them an excellent testbed for transfer learning. Comprehensive graph statistics, label distributions, and text attribute descriptions are provided in Appendix~\ref{app:datasets}.

\textbf{LODO on DTGB.} To apply the LODO protocol on DTGB, each model undergoes four separate pretraining runs, each excluding the datasets of one domain. For example, the model evaluated on the e-commerce domain is pretrained only on datasets from corporate communication (Enron), technical Q\&A forums (Stack E and Stack U), and political events (GDELT and ICEWS), excluding Amazon, Google, and Yelp. This ensures no data from the target domain is observed during pretraining.

\textbf{Pretraining and Finetuning Procedure.} We implement a chronological interleaving strategy that alternates batches from source datasets in a round-robin sequence (details in Appendix~\ref{app:interleaving}). We follow DTGB in partitioning graphs into chronological 70-15-15 training, validation, and test splits. We cap the total pretraining data at 1M training edges and 150k validation edges distributed evenly across the source graphs. Pretraining is performed until convergence, with early stopping based on validation performance using a patience of 10 epochs. During finetuning, the model is adapted using the latest 500 labeled edges from the training split for a fixed 20 epochs.

\begin{table*}[t]
\centering
\small
\setlength{\tabcolsep}{3.5pt}
\caption{\textbf{Dynamic Edge Classification Results.} Values report average Macro F1 (\%) $\pm$ standard deviation over 10 runs. The best performance in each column is shown in \textbf{bold}. Statistical significance of the best-in-column result over the strongest baseline is indicated by $^{\ast} p < 0.10$, $^{\ast\ast} p < 0.05$, and $^{\ast\ast\ast} p < 0.01$ under a two-tailed paired $t$-test for individual datasets and Wilcoxon signed-rank test for the average. In each method block, the superior variant for each dataset is \underline{underlined}, comparing models with pretraining (\textit{W/ PT}) and without pretraining (\textit{W/O PT}).}
\label{tab:finetune_with_nopt_500}
\resizebox{\textwidth}{!}{%
\begin{tabular}{lcccccccccc}
\toprule
\textbf{Method} & \textbf{Amazon} & \textbf{Enron} & \textbf{GDELT} & \textbf{Google} & \textbf{ICEWS} & \textbf{Stack E} & \textbf{Stack U} & \textbf{Yelp} & \textbf{Avg} \\
\midrule
Majority & 15.76$_{\pm 0.00}$ & 4.06$_{\pm 0.00}$ & 0.06$_{\pm 0.00}$ & 15.93$_{\pm 0.00}$ & 0.12$_{\pm 0.00}$ & 42.82$_{\pm 0.00}$ & 44.65$_{\pm 0.00}$ & 13.99$_{\pm 0.00}$ & 17.17$_{\pm 0.00}$ \\
Random & 15.26$_{\pm 0.04}$ & 7.00$_{\pm 0.06}$ & 0.16$_{\pm 0.01}$ & 15.02$_{\pm 0.06}$ & 0.13$_{\pm 0.01}$ & 46.72$_{\pm 0.09}$ & 44.81$_{\pm 0.10}$ & 16.95$_{\pm 0.03}$ & 18.26$_{\pm 0.05}$ \\
\midrule
\multicolumn{10}{l}{GCAL} \\
\hspace{3mm} \textit{W/O PT} & \underline{15.76$_{\pm 0.00}$} & \underline{4.20$_{\pm 1.60}$} & 0.04$_{\pm 0.03}$ & \underline{15.93$_{\pm 0.00}$} & 0.09$_{\pm 0.04}$ & \underline{42.82$_{\pm 0.00}$} & \underline{44.65$_{\pm 0.00}$} & \underline{13.99$_{\pm 0.00}$} & \underline{17.19$_{\pm 0.21}$} \\
\hspace{3mm} \textit{W/ PT} & \underline{15.76$_{\pm 0.00}$} & 3.59$_{\pm 2.61}$ & \underline{0.05$_{\pm 0.02}$} & \underline{15.93$_{\pm 0.00}$} & \underline{0.10$_{\pm 0.05}$} & \underline{42.82$_{\pm 0.00}$} & \underline{44.65$_{\pm 0.00}$} & \underline{13.99$_{\pm 0.00}$} & 17.11$_{\pm 0.34}$ \\
\multicolumn{10}{l}{IDOL} \\
\hspace{3mm} \textit{W/O PT} & \underline{15.76$_{\pm 0.00}$} & 4.06$_{\pm 0.01}$ & 0.12$_{\pm 0.03}$ & \underline{15.93$_{\pm 0.00}$} & \underline{0.13$_{\pm 0.04}$} & \underline{42.82$_{\pm 0.00}$} & \underline{44.65$_{\pm 0.00}$} & \underline{13.99$_{\pm 0.00}$} & 17.18$_{\pm 0.01}$ \\
\hspace{3mm} \textit{W/ PT} & \underline{15.76$_{\pm 0.00}$} & \underline{4.07$_{\pm 0.02}$} & \underline{0.12$_{\pm 0.03}$} & \underline{15.93$_{\pm 0.00}$} & 0.13$_{\pm 0.02}$ & \underline{42.82$_{\pm 0.00}$} & \underline{44.65$_{\pm 0.00}$} & \underline{13.99$_{\pm 0.00}$} & \underline{17.18$_{\pm 0.01}$} \\
\multicolumn{10}{l}{DyGMAE} \\
\hspace{3mm} \textit{W/O PT} & \underline{15.76$_{\pm 0.00}$} & 4.34$_{\pm 0.88}$ & 0.05$_{\pm 0.03}$ & \underline{15.93$_{\pm 0.00}$} & 0.08$_{\pm 0.04}$ & \underline{42.82$_{\pm 0.00}$} & \underline{44.65$_{\pm 0.00}$} & \underline{13.99$_{\pm 0.00}$} & 17.20$_{\pm 0.12}$ \\
\hspace{3mm} \textit{W/ PT} & \underline{15.76$_{\pm 0.00}$} & \underline{4.95$_{\pm 1.75}$} & \underline{0.07$_{\pm 0.03}$} & \underline{15.93$_{\pm 0.00}$} & \underline{0.09$_{\pm 0.05}$} & \underline{42.82$_{\pm 0.00}$} & \underline{44.65$_{\pm 0.00}$} & \underline{13.99$_{\pm 0.00}$} & \underline{17.28$_{\pm 0.23}$} \\
\multicolumn{10}{l}{TLP} \\
\hspace{3mm} \textit{W/O PT} & \underline{15.76$_{\pm 0.00}$} & 4.10$_{\pm 1.26}$ & 0.04$_{\pm 0.02}$ & 15.93$_{\pm 0.00}$ & \underline{0.09$_{\pm 0.04}$} & \underline{42.82$_{\pm 0.00}$} & \underline{44.65$_{\pm 0.00}$} & \underline{13.99$_{\pm 0.00}$} & 17.17$_{\pm 0.17}$ \\
\hspace{3mm} \textit{W/ PT} & \underline{15.76$_{\pm 0.00}$} & \underline{5.37$_{\pm 2.12}$} & \underline{0.06$_{\pm 0.02}$} & \underline{15.94$_{\pm 0.00}$} & 0.08$_{\pm 0.04}$ & \underline{42.82$_{\pm 0.00}$} & \underline{44.65$_{\pm 0.00}$} & \underline{13.99$_{\pm 0.00}$} & \underline{17.33$_{\pm 0.27}$} \\
\multicolumn{10}{l}{DVGMAE} \\
\hspace{3mm} \textit{W/O PT} & 11.91$_{\pm 5.93}$ & 3.84$_{\pm 2.46}$ & 0.02$_{\pm 0.02}$ & 11.45$_{\pm 6.16}$ & 0.02$_{\pm 0.03}$ & 38.45$_{\pm 9.22}$ & 39.35$_{\pm 11.20}$ & 9.37$_{\pm 5.31}$ & 14.30$_{\pm 5.04}$ \\
\hspace{3mm} \textit{W/ PT} & \underline{15.76$_{\pm 0.00}$} & \underline{5.66$_{\pm 2.07}$} & \underline{0.05$_{\pm 0.03}$} & \underline{15.93$_{\pm 0.00}$} & \underline{0.07$_{\pm 0.05}$} & \underline{42.82$_{\pm 0.00}$} & \underline{44.65$_{\pm 0.00}$} & \underline{13.99$_{\pm 0.00}$} & \underline{17.37$_{\pm 0.27}$} \\
\multicolumn{10}{l}{CLDG} \\
\hspace{3mm} \textit{W/O PT} & 15.76$_{\pm 0.00}$ & 5.64$_{\pm 1.71}$ & 0.04$_{\pm 0.02}$ & 15.93$_{\pm 0.00}$ & 0.05$_{\pm 0.03}$ & \underline{42.82$_{\pm 0.00}$} & \underline{44.65$_{\pm 0.00}$} & 13.99$_{\pm 0.00}$ & 17.36$_{\pm 0.22}$ \\
\hspace{3mm} \textit{W/ PT} & \underline{15.76$_{\pm 0.01}$} & \underline{5.94$_{\pm 2.08}$} & \underline{0.06$_{\pm 0.04}$} & \underline{15.94$_{\pm 0.01}$} & \underline{0.07$_{\pm 0.03}$} & \underline{42.82$_{\pm 0.00}$} & \underline{44.65$_{\pm 0.00}$} & \underline{13.99$_{\pm 0.00}$} & \underline{17.40$_{\pm 0.27}$} \\
\multicolumn{10}{l}{EvoluNet} \\
\hspace{3mm} \textit{W/O PT} & 13.44$_{\pm 5.19}$ & 4.19$_{\pm 2.03}$ & 0.02$_{\pm 0.02}$ & 12.78$_{\pm 5.84}$ & 0.02$_{\pm 0.02}$ & 38.29$_{\pm 9.54}$ & 38.96$_{\pm 11.99}$ & 12.57$_{\pm 4.38}$ & 15.03$_{\pm 4.88}$ \\
\hspace{3mm} \textit{W/ PT} & \underline{16.15$_{\pm 0.46}$} & \underline{5.18$_{\pm 1.98}$} & \underline{0.04$_{\pm 0.03}$} & \underline{16.06$_{\pm 0.38}$} & \underline{0.06$_{\pm 0.04}$} & \underline{44.41$_{\pm 2.15}$} & \underline{45.58$_{\pm 1.43}$} & \underline{14.61$_{\pm 0.56}$} & \underline{17.76$_{\pm 0.88}$} \\
\multicolumn{10}{l}{CrossLink} \\
\hspace{3mm} \textit{W/O PT} & \underline{15.76$_{\pm 0.00}$} & 5.27$_{\pm 0.68}$ & 0.08$_{\pm 0.04}$ & \underline{15.93$_{\pm 0.00}$} & \underline{0.12$_{\pm 0.00}$} & 42.82$_{\pm 0.00}$ & 44.65$_{\pm 0.00}$ & 17.28$_{\pm 2.24}$ & 17.74$_{\pm 0.37}$ \\
\hspace{3mm} \textit{W/ PT} & \underline{15.76$_{\pm 0.00}$} & \underline{6.65$_{\pm 1.11}$} & \underline{0.09$_{\pm 0.04}$} & \underline{15.93$_{\pm 0.00}$} & 0.12$_{\pm 0.03}$ & \underline{42.97$_{\pm 0.17}$} & \underline{49.19$_{\pm 0.85}$} & \underline{18.47$_{\pm 1.43}$} & \underline{18.65$_{\pm 0.46}$} \\
\multicolumn{10}{l}{DyGnROLE} \\
\hspace{3mm} \textit{W/O PT} & 15.76$_{\pm 0.00}$ & 4.14$_{\pm 0.14}$ & 0.07$_{\pm 0.01}$ & 15.93$_{\pm 0.00}$ & 0.12$_{\pm 0.00}$ & 42.82$_{\pm 0.00}$ & 44.65$_{\pm 0.00}$ & 13.99$_{\pm 0.00}$ & 17.18$_{\pm 0.02}$ \\
\hspace{3mm} \textit{W/ PT} & \underline{17.06$_{\pm 0.30}$} & \underline{8.67$_{\pm 0.54}$} & \underline{0.27$_{\pm 0.02}$} & \underline{16.13$_{\pm 0.24}$} & \underline{0.43$_{\pm 0.08}$} & \underline{47.56$_{\pm 3.13}$} & \underline{46.54$_{\pm 0.97}$} & \underline{16.94$_{\pm 1.40}$} & \underline{19.20$_{\pm 0.83}$} \\
\multicolumn{10}{l}{BoE} \\
\hspace{3mm} \textit{W/O PT} & 17.44$_{\pm 1.55}$ & \underline{7.78$_{\pm 0.18}$} & 0.15$_{\pm 0.05}$ & 15.95$_{\pm 0.05}$ & 0.16$_{\pm 0.05}$ & 45.79$_{\pm 3.84}$ & 45.28$_{\pm 0.91}$ & 16.80$_{\pm 4.00}$ & 18.67$_{\pm 1.33}$ \\
\hspace{3mm} \textit{W/ PT} & \underline{24.61$_{\pm 0.86}$} & 7.71$_{\pm 0.38}$ & \underline{0.28$_{\pm 0.03}$} & \underline{18.11$_{\pm 1.16}$} & \underline{0.54$_{\pm 0.11}$} & \underline{54.22$_{\pm 1.34}$} & \underline{51.61$_{\pm 1.27}$} & \underline{26.34$_{\pm 2.34}$} & \underline{22.93$_{\pm 0.94}$} \\
\multicolumn{10}{l}{STSA} \\
\hspace{3mm} \textit{W/O PT} & 16.68$_{\pm 1.75}$ & 7.95$_{\pm 0.31}$ & 0.17$_{\pm 0.02}$ & 16.51$_{\pm 0.67}$ & 0.21$_{\pm 0.03}$ & 46.07$_{\pm 2.54}$ & 45.44$_{\pm 1.14}$ & 19.77$_{\pm 3.21}$ & 19.10$_{\pm 1.21}$ \\
\hspace{3mm} \textit{W/ PT} & \underline{\textbf{25.22}$_{\pm 0.40}$$^{\ast}$} & \underline{\textbf{9.12}$_{\pm 0.23}$$^{\ast\ast}$} & \underline{\textbf{0.46}$_{\pm 0.02}$$^{\ast\ast\ast}$} & \underline{\textbf{20.40}$_{\pm 0.72}$$^{\ast\ast\ast}$} & \underline{\textbf{1.24}$_{\pm 0.08}$$^{\ast\ast\ast}$} & \underline{\textbf{55.52}$_{\pm 0.55}$$^{\ast\ast}$} & \underline{\textbf{52.84}$_{\pm 0.76}$$^{\ast\ast}$} & \underline{\textbf{28.96}$_{\pm 0.75}$$^{\ast\ast\ast}$} & \underline{\textbf{24.22}$_{\pm 0.44}$$^{\ast\ast\ast}$} \\
\bottomrule
\end{tabular}%
}
\end{table*}

\textbf{Baselines and Implementation.} We compare BoE and STSA alongside diverse state-of-the-art methods. Our selection includes recent self-supervised methods by \citet{liu2025dygmae}, \citet{gao2025dvgmae}, \citet{xu2023cldg}, \citet{zhu2024idol}, and \citet{bonnet2026dygnrole}, as well as transfer learning methods by \citet{wang2023evolunet}, \citet{huang2025crosslink}, \citet{qiao2025gcal}, and \citet{chatterjee2025tlp}. Implementation details, source code links (where available), and hyperparameter settings are provided in Appendix~\ref{app:hyperparameters}. Hardware specifications are provided in Appendix~\ref{app:compute}.

\subsection{Main Results}
\label{sec:main_results}

\textbf{State-of-the-Art Comparison.} Table~\ref{tab:finetune_with_nopt_500} reveals that existing models struggle to transfer effectively to unseen domains. STSA demonstrates superior performance, consistently achieving substantial performance gains. It also yields the highest average positive transfer (5.12\%), defined as the performance difference between models with and without pretraining. While nearly all models achieve some positive transfer on average, over half collapse to majority class predictors or degrade below random chance on multiple datasets. This widespread degradation suggests that existing mechanisms (e.g., continual memory replay in GCAL, node-level temporal consistency in IDOL) anchor too rigidly to source-domain topologies and cause interference when transferring to target domains.

\textbf{Vulnerabilities of Existing Methods.} Perhaps most strikingly, the minimalist BoE baseline outperforms all existing state-of-the-art methods. Despite being structurally and temporally unaware, operating solely on unordered sequences of historical text features, it achieves an average Macro F1 score of 22.93\%, surpassing the next best model, DyGnROLE, by 3.73 points. This exposes a critical vulnerability in current methods: they overfit to the domain-specific structural, temporal, and linguistic patterns of their pretraining data, and consequently their learned representations fail to provide transferable discriminative power when moved to disjoint domain distributions.

\textbf{Impact of Label Scarcity.} Many DTGB datasets suffer from severe class imbalance (see Figure~\ref{fig:dtgb_label_distributions} in Appendix~\ref{app:datasets}). We observed a strong, statistically significant positive correlation ($r_s = +0.905, p = 0.002$) between training split class imbalance (Gini coefficient) and the relative performance gain of STSA over the BoE baseline, demonstrating that the more a dataset relies on transfer learning due to label scarcity, the greater the performance improvement. In this setting, STSA outperforms BoE by relative margins of 64.29\% on GDELT and 129.63\% on ICEWS. The low absolute scores on GDELT and ICEWS occur because 147 and 169 classes, respectively, are entirely absent from the finetuning split, severely diluting the unweighted Macro F1. Evaluating solely on observed classes significantly expands the absolute margins of STSA over all baselines (Appendix~\ref{app:unseen_classes}).

\begin{figure}[htbp]
    \centering
    \includegraphics[width=0.7\textwidth]{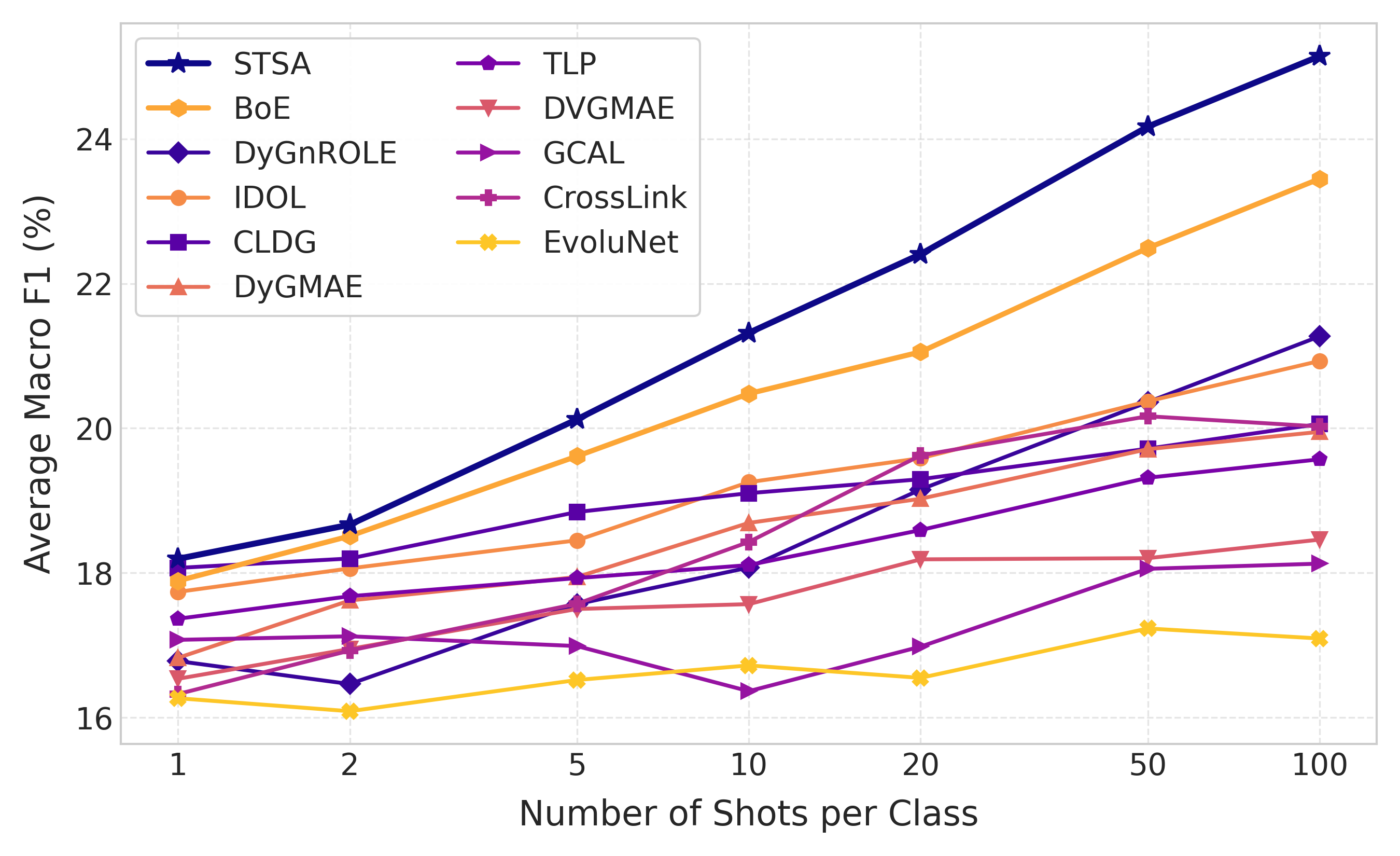}
    \caption{\textbf{Parametric K-Shot Linear Probe Performance.} Average Macro F1 across all target datasets under the leave-one-domain-out (LODO) protocol using a frozen pretrained backbone and a Logistic Regression classifier fitted on $K$ shots per class.}
    \label{fig:linear_probe}
\end{figure}

\subsection{Linear Probing of Frozen Representations}
\label{sec:linear_probe}

While our main results evaluate models under a finetuning paradigm, end-to-end adaptation can obscure the inherent quality of the pretrained representations. To rigorously evaluate the linear separability of the transferred representations without the confounding factor of backbone adaptation, we conduct a parametric $K$-shot linear probing experiment.

In this setup, we freeze the pretrained encoders, and for each target graph, we extract node representations and fit a simple $\ell_2$-regularized multinomial Logistic Regression classifier. To mitigate the effects of class imbalance and evaluate label efficiency, we randomly sample $K$ instances per class from the target training split to train the classifier, where $K \in \{1, 2, 5, 10, 20, 50, 100\}$. The model is then evaluated on the full test split. 

The results, illustrated in Figure~\ref{fig:linear_probe}, reinforce the findings of our finetuning evaluation while exposing weaknesses in existing methods. STSA consistently achieves superior performance across $K$. As the number of examples increases, the performance gap between STSA and the other models widens.

\subsection{Ablation Study}
\label{sec:ablation}

In this section, we evaluate the contribution of the STE module and CSF pretraining objective in our STSA baseline model (Table~\ref{tab:clean_ablation_finetune_500_csf_ste}).

\begin{table*}[t]
\centering
\small
\caption{\textbf{Ablation study of STSA finetuning for dynamic edge classification.} Values report average Macro F1 (\%) $\pm$ standard deviation for individual datasets over 10 runs with different random seeds. Best performance in each column is shown in \textbf{bold}.}
\label{tab:clean_ablation_finetune_500_csf_ste}
\setlength{\tabcolsep}{3.5pt}
\resizebox{\textwidth}{!}{%
\begin{tabular}{lcccccccccc}
\toprule
\textbf{Variant} & \textbf{Amazon} & \textbf{Enron} & \textbf{GDELT} & \textbf{Google} & \textbf{ICEWS} & \textbf{Stack E} & \textbf{Stack U} & \textbf{Yelp} & \textbf{Avg} \\
\midrule
\multicolumn{10}{l}{\textbf{\textit{Reference Baseline}}} \\
\midrule
BoE & 24.61$_{\pm 0.86}$ & 7.71$_{\pm 0.38}$ & 0.28$_{\pm 0.03}$ & 18.11$_{\pm 1.16}$ & 0.54$_{\pm 0.11}$ & 54.22$_{\pm 1.34}$ & 51.61$_{\pm 1.27}$ & 26.34$_{\pm 2.34}$ & 22.93$_{\pm 0.94}$ \\
\midrule
\multicolumn{10}{l}{\textbf{\textit{STSA Variants}}} \\
\midrule
w/o STE & 24.14$_{\pm 0.54}$ & 8.18$_{\pm 0.38}$ & 0.31$_{\pm 0.02}$ & 18.07$_{\pm 0.56}$ & 0.70$_{\pm 0.09}$ & 51.76$_{\pm 2.69}$ & 51.16$_{\pm 1.72}$ & 28.16$_{\pm 0.53}$ & 22.81$_{\pm 0.82}$ \\
CSF $\to$ SF & 23.79$_{\pm 0.47}$ & 8.34$_{\pm 0.42}$ & 0.41$_{\pm 0.02}$ & 20.23$_{\pm 0.72}$ & 1.12$_{\pm 0.06}$ & 54.38$_{\pm 2.29}$ & 52.71$_{\pm 0.78}$ & 28.16$_{\pm 0.38}$ & 23.64$_{\pm 0.64}$ \\
CSF $\to$ InfoNCE & 21.72$_{\pm 0.62}$ & \textbf{10.53}$_{\pm 1.21}$ & 0.37$_{\pm 0.01}$ & 18.47$_{\pm 0.79}$ & 0.95$_{\pm 0.06}$ & 54.08$_{\pm 1.22}$ & 50.88$_{\pm 0.94}$ & 26.97$_{\pm 0.54}$ & 22.99$_{\pm 0.68}$ \\
\midrule
Complete STSA & \textbf{25.22}$_{\pm 0.40}$ & 9.12$_{\pm 0.23}$ & \textbf{0.46}$_{\pm 0.02}$ & \textbf{20.40}$_{\pm 0.72}$ & \textbf{1.24}$_{\pm 0.08}$ & \textbf{55.52}$_{\pm 0.55}$ & \textbf{52.84}$_{\pm 0.76}$ & \textbf{28.96}$_{\pm 0.75}$ & \textbf{24.22}$_{\pm 0.44}$ \\
\bottomrule
\end{tabular}%
}
\end{table*}

\textbf{STE.} Bypassing the STE and feeding concatenated structural and temporal features directly into the Transformer (w/o STE) reduces the average Macro F1 from 24.22 to 22.81. This confirms the findings of our representational analysis: non-linear fusion via the STE is necessary to mitigate representational fragmentation and prevent a drop in eRank when transferring to unseen domains.

\textbf{CSF.} Swapping the CSF objective for a standard InfoNCE loss (CSF $\to$ InfoNCE) results in a drop in average Macro F1 from 24.22 to 22.99, exposing the limitations of purely structural objectives that ignore query edge semantics. However, a notable exception occurs on the Enron dataset, where InfoNCE outperforms CSF. This suggests that in networks defined by organizational cliques, the structural identity of the dyad can act as a stronger predictor of the interaction topic than the semantic trajectory of past interactions. Crucially, this structural alignment only succeeds in combination with the STE. InfoNCE without the STE (the BoE baseline) falls to 7.71. Finally, replacing CSF with a direct semantic forecasting objective using an SCE \citep{hou2022graphmae} loss (CSF $\to$ SF) degrades average performance to 23.64, demonstrating that contrastive alignment yields a more robust transfer prior than point-by-point vector reconstruction. We provide the formulation of SCE in Appendix~\ref{app:pretraining_objectives}.

\section{Conclusion}
\label{conclusion}

We formally established a LODO protocol to investigate transfer learning for edge classification on DyTAGs. Our evaluations exposed a critical vulnerability in current dynamic graph methods: they overfit to the structural, temporal, and semantic distributions of their pretraining domains. Consequently, complex models fail to generalize to unseen networks, consistently underperforming our structurally and temporally unaware BoE baseline. To address this, we proposed STSA, a framework that uses a spatio-temporal encoder to non-linearly fuse temporal and structural representations, mitigating fragmentation and expanding effective capacity. This is paired with a CSF objective that anchors edge representations to a multi-domain PLM latent space. STSA demonstrates superior performance and linear separability across all evaluated DTGB domains. Ultimately, this work underscores the importance of evaluating models under severe distribution shifts and provides a robust foundation for learning generalizable interaction patterns.

\subsection*{AI use statement}

Generative AI was used to improve text readability.

\subsection*{Ethics statement}

This work focuses on foundational methodology for dynamic graph forecasting. However, predicting future interactions in networks, such as political events or corporate communications, presents dual-use risks. In downstream deployments, such predictive models could be used for unwarranted surveillance of individuals or groups. Mitigating these risks in applied settings requires careful consideration of the deployment context and the implementation of appropriate access controls when using human-centric datasets.

\subsection*{Reproducibility statement}

We detail the leave-one-domain-out protocol in Section~\ref{sec:setup} and provide all hyperparameters in Appendix~\ref{app:hyperparameters}. The datasets are accessible via the DTGB repository (\url{https://github.com/zjs123/DTGB}), and code will be made public upon publication. Detailed architectural descriptions and hyperparameter settings in the main text and appendices support reproducibility in the interim.

\subsubsection*{Acknowledgments}

We thank Xiaowen Dong (University of Oxford) for his helpful feedback.

\bibliography{iclr2027_conference}
\bibliographystyle{iclr2027_conference}

\appendix

\section{BoE Architecture Illustration}
\label{app:boe_architecture}

\begin{figure}[htbp]
    \centering
    \includegraphics[width=\textwidth]{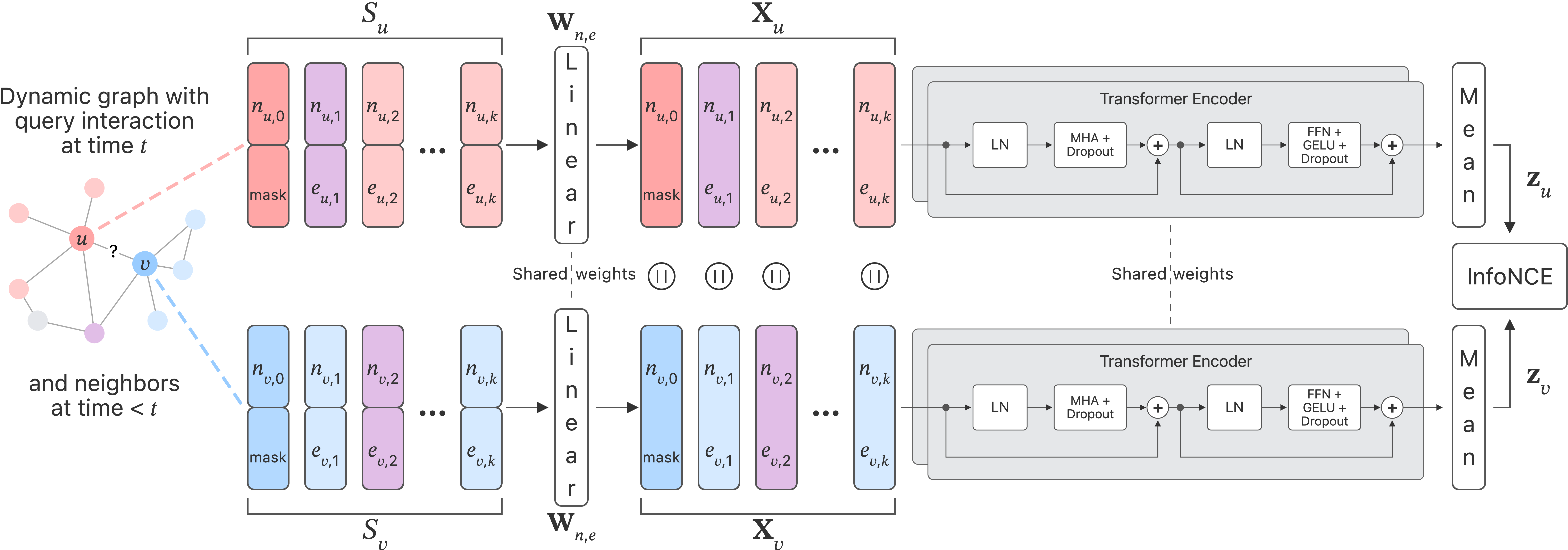}
    \caption{\textbf{Overview of the Bag of Events (BoE) Baseline.} \textbf{(Left)} For a query interaction between nodes $u$ and $v$ at time $t$, historical one-hop neighbors are retrieved. $u$ and $v$ combined with their respective neighbors form node sequences $S_u$ and $S_v$. For each event $i$ in these sequences, its node text embedding $\mathbf{n}_i$ and edge text embedding $\mathbf{e}_i$ are extracted, and the edge text embedding of the query nodes is masked. \textbf{(Center)} Both extracted features $\mathbf{n}_i$ and $\mathbf{e}_i$ are first independently projected via separate linear layers $\mathbf{W}_{n,e}$. These projected embeddings are then concatenated to form a single feature vector representing each event in $S_u$ and $S_v$. \textbf{(Right)} These event sequences, corresponding to input matrices $\mathbf{X}_u$ and $\mathbf{X}_v$, are processed independently by a multi-layer Transformer encoder. Mean pooling, with a mask to exclude padding tokens, is applied to form the final node representations $\mathbf{z}_u$ and $\mathbf{z}_v$. These representations are optimized via a symmetric InfoNCE loss.
}

    \label{fig:boe_architecture}
\end{figure}

\section{Self-Supervised Pretraining Objectives}
\label{app:pretraining_objectives}

This section provides the formal mathematical formulations for the self-supervised pretraining objectives evaluated in this work, specifically the symmetric InfoNCE loss used for the Bag of Events (BoE) baseline and the Scaled Cosine Error (SCE) used in our ablation study.

\paragraph{Symmetric InfoNCE Loss}
For the BoE baseline, the model optimizes a symmetric contrastive objective to align the independent sequence representations of interacting nodes $u$ and $v$. Given a minibatch of size $N$, let $\mathbf{z}_u^{(i)}$ and $\mathbf{z}_v^{(i)}$ denote the $L_2$-normalized, pooled representations for the $i$-th interaction in the batch. The symmetric InfoNCE loss \citep{oord2018infonce} is formulated as:

\begin{equation}
    \label{eq:infonce}
    \mathcal{L}_{\mathrm{InfoNCE}} = -\frac{1}{2N} \sum_{i=1}^{N} \left( \log \frac{\exp(\mathbf{z}_u^{(i)} \cdot \mathbf{z}_v^{(i)} / \tau)}{\sum_{j=1}^{N} \exp(\mathbf{z}_u^{(i)} \cdot \mathbf{z}_v^{(j)} / \tau)} + \log \frac{\exp(\mathbf{z}_v^{(i)} \cdot \mathbf{z}_u^{(i)} / \tau)}{\sum_{j=1}^{N} \exp(\mathbf{z}_v^{(i)} \cdot \mathbf{z}_u^{(j)} / \tau)} \right),
\end{equation}

where $\tau$ is a temperature hyperparameter set to $0.07$. The first term matches the representation of node $u$ with its true interacting partner $v$ against all other in-batch negative representations, while the second term performs the symmetric matching for node $v$.

\paragraph{Scaled Cosine Error (SCE)}
In the ablation study (Section~\ref{sec:ablation}), we evaluate a direct semantic forecasting (SF) variant of our model that replaces the CSF objective with a vector reconstruction loss, SCE, proposed by \citet{hou2022graphmae}. Let $\mathbf{z}_{pred}^{(i)}$ be the predicted representation and $\mathbf{z}_{target}^{(i)}$ be the true edge text embedding. The SCE loss is defined as:

\begin{equation}
    \label{eq:sce}
    \mathcal{L}_{\mathrm{SCE}} = \frac{1}{N} \sum_{i=1}^{N} \left( 1 - \frac{\mathbf{z}_{pred}^{(i)} \cdot \mathbf{z}_{target}^{(i)}}{\|\mathbf{z}_{pred}^{(i)}\|_2 \|\mathbf{z}_{target}^{(i)}\|_2} \right)^\gamma,
\end{equation}

where $\gamma$ is a scaling factor that down-weights easily reconstructed samples, incentivizing the model to focus training capacity on interaction dynamics that are more difficult to predict.

\section{Synthetic Graph Generation Parameters}
\label{app:synthetic_generation}

For the representational analysis detailed in Section~\ref{sec:rep_analysis}, we generate synthetic dynamic graphs for four distinct archetypes: Non-Recurrent Bipartite, Recurrent Bipartite, Asymmetric Homogeneous, and Symmetric Homogeneous. These archetypes simulate distinct structural patterns: one-off user-item reviews, repeated user-post interactions, imbalanced communication networks, and highly reciprocal event networks, respectively. Each graph contains 60 nodes and 1,500 interactions. We control structural variance across topologies by altering power-law degree distributions, community boundaries, node recurrence probabilities, and edge reciprocity. We control temporal variance by sampling interaction intervals from exponential distributions with varying scale parameters to produce both dense temporal bursts and sparse activity. Edges are partitioned into an 80\% pretraining split and a 20\% validation split for the InfoNCE objective. Below are the generation parameters.

\textbf{Non-Recurrent Bipartite.} Nodes are partitioned into 36 sources and 24 destinations. Source selection probability is proportional to $i^{-0.5}$, and destination probability to $i^{-0.8}$. Recurrence is strictly prohibited until a source node exhausts all 24 available destinations, at which point its history resets. This finite pool exhaustion by high-activity nodes creates a marginal fraction (1.48\%) of recurrent edges where the root node appears in the destination sequence, producing $(w=1, c=1)$  tokens. Time steps $\Delta t$ between interactions are drawn from $\mathrm{Exp}(5.0) + 1$, producing slow, sparse activity.

\textbf{Recurrent Bipartite.} Nodes are equally partitioned into 30 sources and 30 destinations. Source selection probability is proportional to $i^{-1.1}$. There is a 45\% probability of recurrence. Time steps are fast, drawn from $\mathrm{Exp}(0.4)$ plus an additional step of 1 with a 60\% probability.

\textbf{Asymmetric Homogeneous.} Nodes are divided into four 15-node communities. Out-degree probability is proportional to $i^{-0.9}$. In-degree probability is a random permutation of the out-degree distribution, breaking symmetry. We apply a 75\% intra-community bias for destination selection. Time increments are binary (1 with a 15\% probability, 0 otherwise) to simulate simultaneous interactions. Additionally, there is a 10\% probability of immediate edge reciprocity ($v \rightarrow u$).

\textbf{Symmetric Homogeneous.} Similar to the asymmetric variant, nodes form four 15-node communities with out-degree probability proportional to $i^{-0.9}$. However, in-degree weights exactly match out-degree weights. The intra-community selection bias is increased to 80\%. Time increments are 1 with a 12\% probability. The immediate edge reciprocity rate is heavily increased to 55\%, simulating dense, highly communicative subgraphs. 

\section{Spatio-Temporal Representations Colored by Node Frequency}
\label{app:ste_geometry_node_frequency}

\begin{figure}[H]
    \centering
    \includegraphics[width=\textwidth]{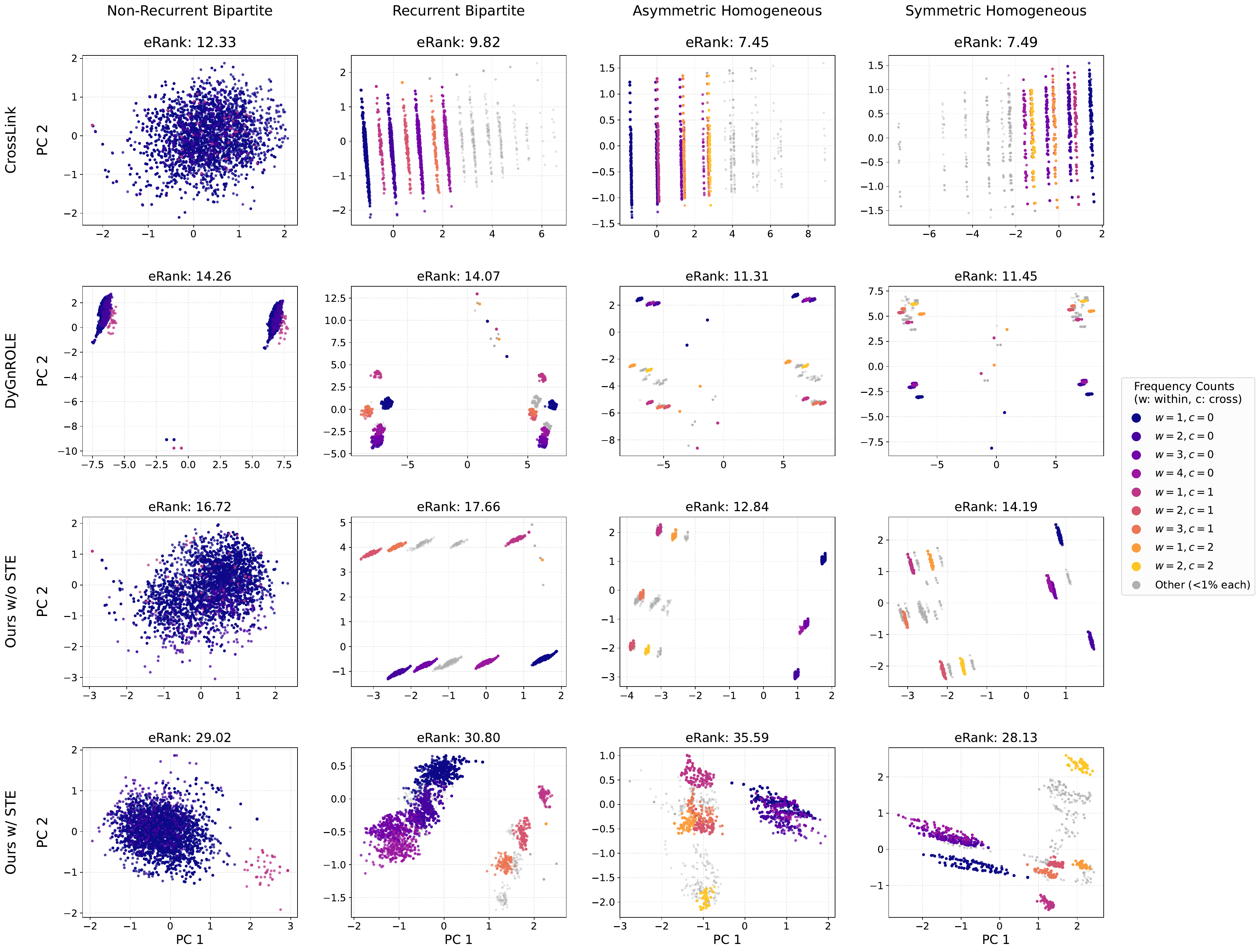}
    \caption{\textbf{Analysis of Spatio-Temporal Representations.} Principal Component Analysis (PCA) of spatio-temporal representations colored by combinations of within- and cross-sequence node frequency counts across four synthetic graph topologies. \textbf{Top 3}: Under domain transfer, concatenated time and node frequency features suffer from representational fragmentation and a restricted Effective Rank (eRank). \textbf{Bottom}: STE fuses the temporal and structural features into a unified manifold and substantially improves eRank.}
    \label{fig:ste_geometry_node_frequency}
\end{figure}

\section{Dataset Details}
\label{app:datasets}

This section provides details for the datasets from the Dynamic Text-Attributed Graph Benchmark \citep{zhang2024dtgb} used in our evaluation, including summary statistics in Table~\ref{tab:dtgb_stats_table}. The underlying raw data for these benchmarks originates from the publicly available sources cited below.

\begin{table}[h]
    \centering
    \small
    \caption{Summary statistics for the DTGB datasets.}
    \label{tab:dtgb_stats_table}
    \begin{tabular}{lrrrl}
    \toprule
    Dataset & \#Nodes & \#Edges & \#Classes & Domain \\
    \midrule
    Enron & 42,711 & 797,907 & 10 & Corporate communication \\
    GDELT & 6,786 & 1,339,245 & 236 & Political events \\
    ICEWS1819 & 31,796 & 1,100,071 & 266 & Political events \\
    Stack elec & 397,702 & 1,262,225 & 2 & Technical Q\&A forum \\
    Stack ubuntu & 674,248 & 1,497,006 & 2 & Technical Q\&A forum \\
    Amazon movies & 293,566 & 3,217,324 & 5 & E-commerce reviews \\
    Googlemap CT & 111,168 & 1,380,623 & 5 & E-commerce reviews \\
    Yelp & 2,138,242 & 6,990,189 & 5 & E-commerce reviews \\
    \bottomrule
    \end{tabular}
\end{table}

\textbf{Enron}\footnote{\url{https://www.cs.cmu.edu/~enron/}} represents a network of email communications between employees of the Enron corporation from 1999 to 2002.
\begin{itemize}
    \item \textbf{Nodes}: Employees
    \item \textbf{Node texts}: Professional email addresses
    \item \textbf{Edges}: Email exchanges between employees
    \item \textbf{Edge texts}: Raw email content
    \item \textbf{Edge labels}: 10 distinct communication categories (e.g., 'deal', 'calendar')
\end{itemize}

\textbf{GDELT}\footnote{\url{https://www.gdeltproject.org/}} is derived from the Global Database of Events, Language, and Tone. It records international political events and interactions.
\begin{itemize}
    \item \textbf{Nodes}: Political entities or actors
    \item \textbf{Node texts}: Entity names
    \item \textbf{Edges}: Relationships or actions between two political entities
    \item \textbf{Edge texts}: Concise descriptions of the relationship
    \item \textbf{Edge labels}: 236 unique relationship types (e.g., 'provide economic aid')
\end{itemize}

\textbf{ICEWS1819 (ICEWS)}\footnote{\url{https://dataverse.harvard.edu/dataverse/icews}} originates from the Integrated Crisis Early Warning System and models political events occurring between 2018 and 2019.
\begin{itemize}
    \item \textbf{Nodes}: Political entities
    \item \textbf{Node texts}: Entity names
    \item \textbf{Edges}: Documented relationships between political actors
    \item \textbf{Edge texts}: Short descriptions of the event or relationship
    \item \textbf{Edge labels}: 266 unique interaction categories (e.g., 'Engage in negotiation')
\end{itemize}

\textbf{Stack elec (Stack E)}\footnote{\url{https://archive.org/details/stackexchange}} consists of data from the Stack Exchange forum focused on electronics and electrical engineering.
\begin{itemize}
    \item \textbf{Nodes}: Users or specific questions
    \item \textbf{Node texts}: User introduction and location: question title and post content
    \item \textbf{Edges}: The submission of an answer or a comment on a post
    \item \textbf{Edge texts}: Textual content of answers or comments
    \item \textbf{Edge labels}: Binary classification ('useful' or 'useless') based on a voting score where $score = upvotes - downvotes$. An edge is 'useful' if $score > 1$.
\end{itemize}

\textbf{Stack ubuntu (Stack U)}\footnote{\url{https://archive.org/details/stackexchange}} follows the same structure as Stack E but focuses on the Ubuntu operating system community.
\begin{itemize}
    \item \textbf{Nodes}: Users or questions
    \item \textbf{Node texts}: User introduction and location: question title and post content
    \item \textbf{Edges}: Answer submissions or comments on existing posts
    \item \textbf{Edge texts}: Textual content of answers or comments
    \item \textbf{Edge labels}: Binary classification ('useful' or 'useless') where $score = upvotes - downvotes$ and 'useful' requires $score > 1$.
\end{itemize}

\textbf{Googlemap CT (Google)}\footnote{\url{https://mcauleylab.ucsd.edu/public_datasets/gdrive/googlelocal/}} consists of business reviews from Google for the state of Connecticut.
\begin{itemize}
    \item \textbf{Nodes}: Users or local businesses
    \item \textbf{Node texts}: Usernames for individuals: business name, address, category, and introduction
    \item \textbf{Edges}: Review interactions between users and businesses
    \item \textbf{Edge texts}: The textual content of user reviews
    \item \textbf{Edge labels}: 1 to 5 based on the standard five-star rating system
\end{itemize}

\textbf{Amazon movies (Amazon)}\footnote{\url{https://cseweb.ucsd.edu/~jmcauley/datasets/amazon_v2/}} consists of review data for the Movies and TV category on Amazon.
\begin{itemize}
    \item \textbf{Nodes}: Users or products in the Movies and TV category
    \item \textbf{Node texts}: Usernames for individuals: product title, description, category, and rank score
    \item \textbf{Edges}: Reviews submitted by users for specific products
    \item \textbf{Edge texts}: The textual content of user reviews
    \item \textbf{Edge labels}: 1 to 5 based on the five-star rating system
\end{itemize}

\textbf{Yelp}\footnote{\url{https://www.yelp.com/dataset}} consists of reviews for businesses such as hotels and restaurants.
\begin{itemize}
    \item \textbf{Nodes}: Users or businesses
    \item \textbf{Node texts}: Name, register date, review count, and average rating for users: business name, city, address, and category
    \item \textbf{Edges}: Review interactions between users and businesses
    \item \textbf{Edge texts}: The textual content of user reviews
    \item \textbf{Edge labels}: 1 to 5 based on the five-star rating system
\end{itemize}

\begin{figure}[tbp]
    \centering
    \includegraphics[width=\columnwidth]{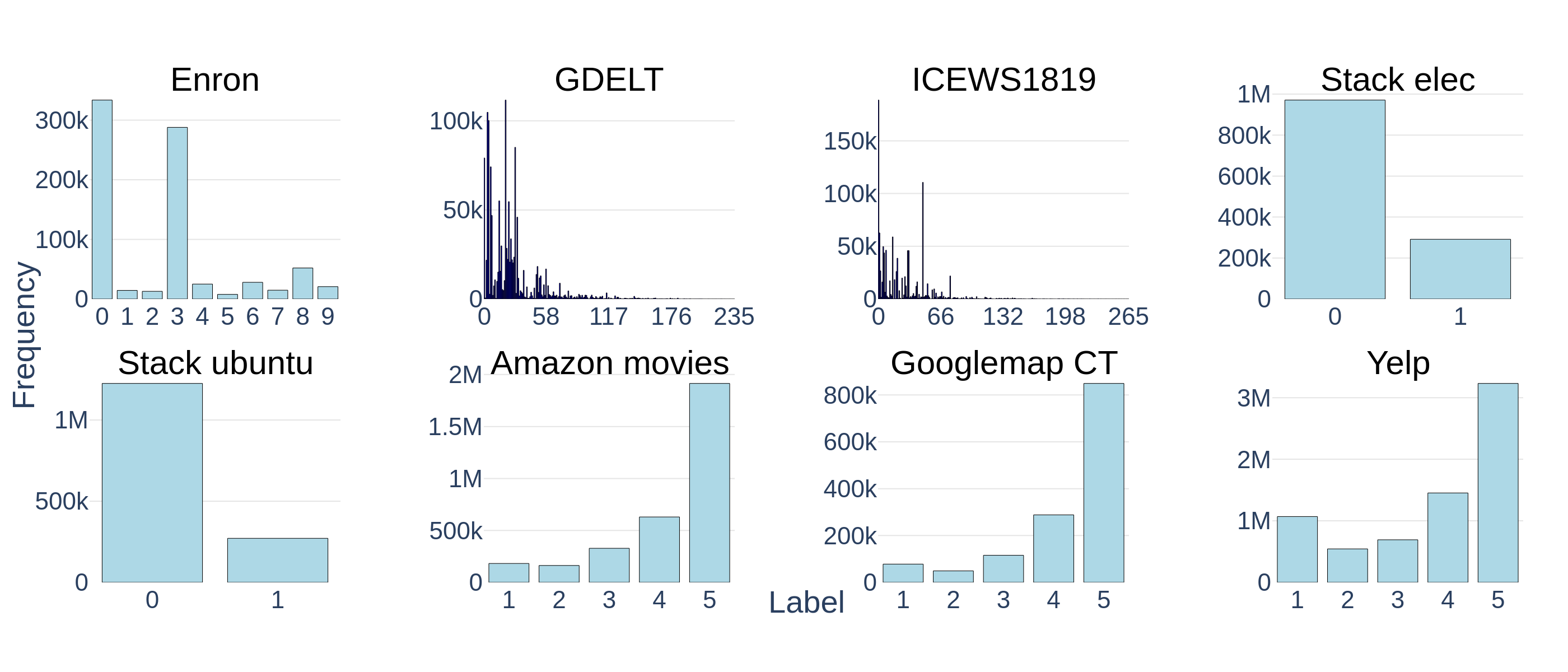}
    \caption{Label distributions of the DTGB datasets.}
    \label{fig:dtgb_label_distributions}
\end{figure}

\section{Multi-Dataset Pretraining and Interleaving Strategy}
\label{app:interleaving}

To facilitate cross-domain pretraining across the heterogeneous datasets in the LODO protocol, we implement a unified data orchestration layer. This layer ensures that the model is exposed to a balanced distribution of interaction dynamics from all source domains simultaneously, preventing the encoder from oscillating between domain-specific local minima.

\paragraph{Feature Stacking via Virtual Offsetting.} Because each dataset in the DTGB benchmark uses an independent coordinate system for node and edge identifiers, naive concatenation leads to collision. We use a virtual feature stacking mechanism that creates a unified global index. For a set of source graphs $\{\mathcal{G}_1, \mathcal{G}_2, \dots, \mathcal{G}_K\}$, we compute a cumulative offset $O_k = \sum_{i=1}^{k-1} |\mathcal{V}_i|$ for nodes and similarly for edges. During pretraining, the raw identifiers in $\mathcal{G}_k$ are mapped to a global space as $ID_{global} = ID_{local} + O_k$. This allows STSA and baseline models to treat the collection of source graphs as a single, large super-graph without requiring expensive physical merges of the feature tensors.
    
\paragraph{Chronological Batch Interleaving.} To facilitate cross-domain pretraining, we use a deterministic round-robin interleaving strategy. While each individual mini-batch contains interactions from a single source dataset to ensure structural consistency, we interleave these batches during training. For $N$ source datasets across the included pretraining domains, the model processes a sequence of $N$ batches, one from each dataset, before repeating the cycle. This ensures that the global gradient trajectory is informed by the collective variance of all source datasets, preventing the encoder from overfitting to the idiosyncratic distribution of any single graph while adhering to the chronological order within each dataset.

\section{Model Hyperparameters}
\label{app:hyperparameters}

We implement all baselines using PyTorch. To ensure a fair comparison, we use standardized pretraining hyperparameters across all models unless a specific architecture requires otherwise.

\textbf{Shared Settings}
\begin{itemize}
    \item Optimizer: AdamW
    \item Learning rate: $1\times 10^{-4}$
    \item Batch size ($N$): 256
    \item Pretraining epochs: 1000
    \item Early stopping patience: 10 epochs
    \item Global random seed: 42
    \item Maximum number of neighbors ($S$): 10
\end{itemize}

\textbf{BoE}
\begin{itemize}
    \item Transformer layers: 2
    \item Transformer heads: 2
    \item Transformer dropout rate: 0.1
    \item Text channel dimension: 96
\end{itemize}

\textbf{STSA}
\begin{itemize}
    \item Transformer layers: 2
    \item Transformer heads: 2
    \item Transformer dropout rate: 0.1
    \item Text channel dimension: 96
    \item Time dimension: 32
    \item Node frequency embedding dimension: 32
    \item Spatio-temporal encoder hidden dimension: 256
    \item Spatio-temporal encoder latent dimension: 96
\end{itemize}

\textbf{IDOL}\footnote{\url{https://github.com/ZulunZhu/dynamic-contrastive-learning}}
\begin{itemize}
    \item Hidden dimension: 512
    \item Embedding dimension: 128
    \item Neighborhood samples ($K$): 4
    \item Temporal similarity weight ($\alpha$): 0.1
    \item Maximum bias ($r_{max}^b$): 0.01
    \item Loss trade-off ($\lambda$): 0.01
\end{itemize}

\textbf{CLDG}\footnote{\url{https://github.com/yimingxu24/CLDG}}
\begin{itemize}
    \item InfoNCE temperature: 0.07
    \item Hidden dimension: 128
    \item Embedding dimension: 64
    \item Encoder layers: 2
    \item Dropout rate: 0.1
\end{itemize}

\textbf{TLP}\footnote{\url{https://github.com/google-research/google-research/tree/master/fm4tlp}}
\begin{itemize}
    \item Alignment weight ($\alpha$): 1.0
    \item Alignment objective: Mean Squared Error
    \item Link prediction objective: Binary Cross-Entropy
    \item Topological normalization: Offline (global statistics)
\end{itemize}

\textbf{GCAL}\footnote{\url{https://github.com/joe817/GCAL}}
\begin{itemize}
    \item Entropy weight: 5.0
    \item Replay weight: 10.0
    \item VAE weight: 1.0
    \item Edge weight: 1.0
    \item Teacher EMA $\alpha$: 0.9
    \item Bilevel optimization inner loops: 10
    \item Memory optimizer learning rate: $1\times 10^{-3}$
    \item Warmup epochs: 2
\end{itemize}

\textbf{DVGMAE}
\begin{itemize}
    \item Hidden dimension: 128
    \item Embedding dimension: 32
    \item Dropout rate: 0.1
    \item Epsilon ($\epsilon$): 0.25
    \item Mask ratio: 0.5
    \item Loss lambda weight ($\lambda$): 0.5
\end{itemize}

\textbf{DyGMAE}
\begin{itemize}
    \item Hidden dimension: 128
    \item Encoder layers: 2
    \item Dropout rate: 0.1
    \item Number of masks: 2
    \item Mask ratios: 0.2, 0.4
    \item Mask types: path, random
    \item Fusion type: attention
    \item Walk length: 2
    \item Loss gamma ($\gamma$): 0.1
    \item Loss delta ($\delta$): 0.1
    \item Loss lambda $c$ ($\lambda_c$): 0.1
\end{itemize}

\textbf{DyGnROLE}
\begin{itemize}
    \item Transformer layers: 2
    \item Transformer heads: 2
    \item Transformer dropout rate: 0.1
    \item Text channel dimension: 32
    \item Time dimension ($d_t$): 32
    \item Node frequency embedding dimension ($d_c$): 32
\end{itemize}

\section{Computing Resources}

\label{app:compute}
Experiments were performed using an Intel Xeon Silver 4214 CPU (2.20GHz), 128 GB of system RAM, four NVIDIA GeForce RTX 2080 Ti GPUs (11 GB VRAM each) for all tasks.

\section{Unseen Class Dilution of Finetuning Results}
\label{app:unseen_classes}

\begin{figure}[htbp]
    \centering
    \includegraphics[width=\textwidth]{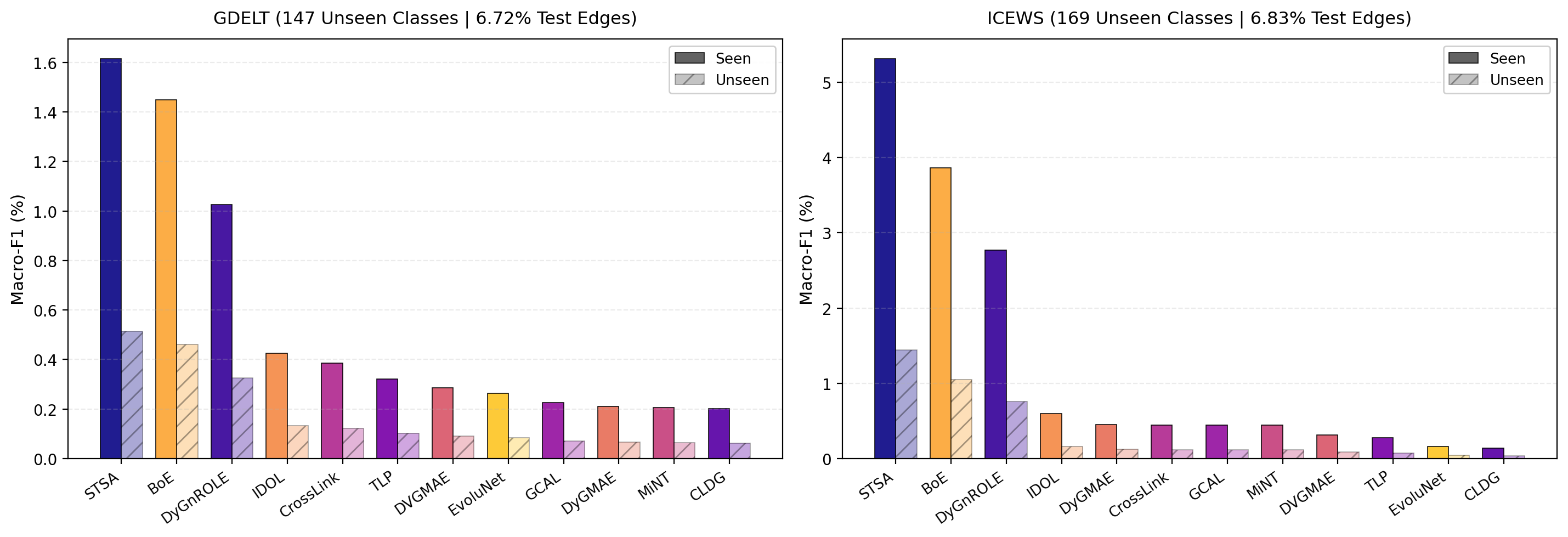}
    \caption{\textbf{Impact of Unseen-Class Dilution on Macro-F1.} Because Macro F1 weights all classes equally, the unobserved classes in the 500 training edges of the finetuning experiment performed in Section~\ref{experiments} drag down the overall score on GDELT and ICEWS. Isolating evaluation to seen classes removes this dilution artifact, revealing larger absolute gains for STSA on GDELT and ICEWS.}
    \label{fig:seen_classes}
\end{figure}

\end{document}